\documentclass[conference]{IEEEtran}
\IEEEoverridecommandlockouts
\usepackage{cite}
\usepackage{amsmath,amssymb,amsfonts}
\usepackage{algorithmic}
\usepackage{graphicx}
\usepackage{textcomp}
\usepackage{xcolor}
\usepackage{subfigure}
\usepackage{url}

\usepackage{amsthm, multirow}
\usepackage{color, colortbl}
\newtheorem{thm}{Theorem}

\def\BibTeX{{\rm B\kern-.05em{\sc i\kern-.025em b}\kern-.08em
    T\kern-.1667em\lower.7ex\hbox{E}\kern-.125emX}}
\begin{document}

\title{Distill-to-Label: Weakly Supervised Instance Labeling Using Knowledge Distillation 
}

 \author{\IEEEauthorblockN{Jayaraman J. Thiagarajan\thanks{This work was performed under the auspices of the U.S. Departmentof Energy by Lawrence Livermore National Laboratory under Contract DE-AC52-07NA27344.}}
 \IEEEauthorblockA{\textit{Lawrence Livermore National Laboratory}\\
 Livermore, USA \\
 jjayaram@llnl.gov}
 \and
 \IEEEauthorblockN{Satyananda Kashyap, Alexandros Karagyris}
 \IEEEauthorblockA{\textit{IBM Research}\\
 Almaden, USA \\
 \{satyananda.kashyap,akararg\}@us.ibm.com}
}

\maketitle

\begin{abstract}
Weakly supervised instance labeling using only image-level labels, in lieu of expensive fine-grained pixel annotations, is crucial in several applications including medical image analysis. In contrast to conventional instance segmentation scenarios in computer vision, the problems that we consider are characterized by a small number of training images and non-local patterns that lead to the diagnosis. In this paper, we explore the use of multiple instance learning (MIL) to design an instance label generator under this weakly supervised setting. Motivated by the observation that an MIL model can handle bags of varying sizes, we propose to repurpose an MIL model originally trained for bag-level classification to produce reliable predictions for single instances, i.e., bags of size $1$. To this end, we introduce a novel regularization strategy based on virtual adversarial training for improving MIL training, and subsequently develop a knowledge distillation technique for repurposing the trained MIL model. Using empirical studies on colon cancer and breast cancer detection from histopathological images, we show that the proposed approach produces high-quality instance-level prediction and significantly outperforms state-of-the MIL methods.
\end{abstract}


\section{Introduction}
The success of supervisory algorithms, e.g. deep neural networks, in computer vision relies on the assumption that labels provide the required level of semantic description for the task at hand. For example, in object recognition~\cite{krizhevsky2012imagenet}, image-level labels are sufficient to identify generalizable patterns for each category, whereas in instance segmentation~\cite{khoreva2017simple,li2017instance}, pixel-level annotations are required to produce detailed segmentation masks. However, in real-world applications, we are required to design effective models under \textit{weakly supervised} settings. This is highly prevalent in medical image analysis~\cite{kandemir2015computer, xu2014weakly, anirudh2016lung}, e.g. computational pathology, wherein an image is typically described by an overall diagnostic label, though patients with similar diagnosis can present vastly different signatures in their diagnostic images. Furthermore, annotating medical images is very expensive due to the required expertise of the clinicians, thus limiting the availability of labeled data~\cite{ratner2017snorkel}.

Consequently, weakly supervised image segmentation methods have gained significant research interest~\cite{zhou2018weakly}, and the most successful approaches typically view filters in convolutional networks as object detectors and aggregate deep feature maps to infer class-aware visual evidence~\cite{li2018tell}. However, medical diagnosis presents unique challenges when compared to conventional object recognition tasks. More specifically, the factors leading to a diagnosis are not well-localized and are based on observing the co-occurrence of seemingly unrelated patterns in different parts of the image, thus making it challenging to infer generalizable features. As a result, recent approaches have explored the use of multiple instance learning (MIL)~\cite{xu2014deep}, wherein each sample is represented as a bag of instances with the assumption that only the bag-level labels are available. At their core, MIL methods attempt to aggregate features from the instances, while exploiting correlations between them, to produce an effective statistical descriptor. Given the complex nature of clinical decision process, more recent MIL methods have also included \textit{interpretability} as an additional design objective. For example, Ilse \textit{et. al.} include a learnable attention module in MIL to automatically identify key instances in a bag that are the most likely to trigger the observed bag-label~\cite{ilse2018attention}.

\noindent \textbf{Proposed Work.} In this paper, we develop a novel approach to generate dense labels for clinical images using only weakly supervised data. To this end, we represent each image as a bag of patches and adopt an MIL formulation for predicting the bag-level diagnostic label. Given the strong dependence of MIL methods on the average number of instances in bags used for training, we propose to incorporate regularization based on \textit{virtual adversarial training} (VAT). In general, VAT optimizes for the network to achieve information invariance under arbitrary perturbations to the input data~\cite{shu2018dirt,ji2018invariant}. The form of perturbation we utilize is designed specifically for MIL -- in addition to corrupting each of the instances, we employ perturbations to the bag, namely pruning instances and including uncorrelated noise instances. We demonstrate that the proposed VAT based regularization leads to highly robust predictive models when compared to existing approaches, particularly when the training data sizes are small. More importantly, we find this regularization to be crucial for improving instance-level predictions of an MIL model.

\begin{figure*}
	\centering
	\subfigure[Teacher model]{\includegraphics[width=0.19\linewidth]{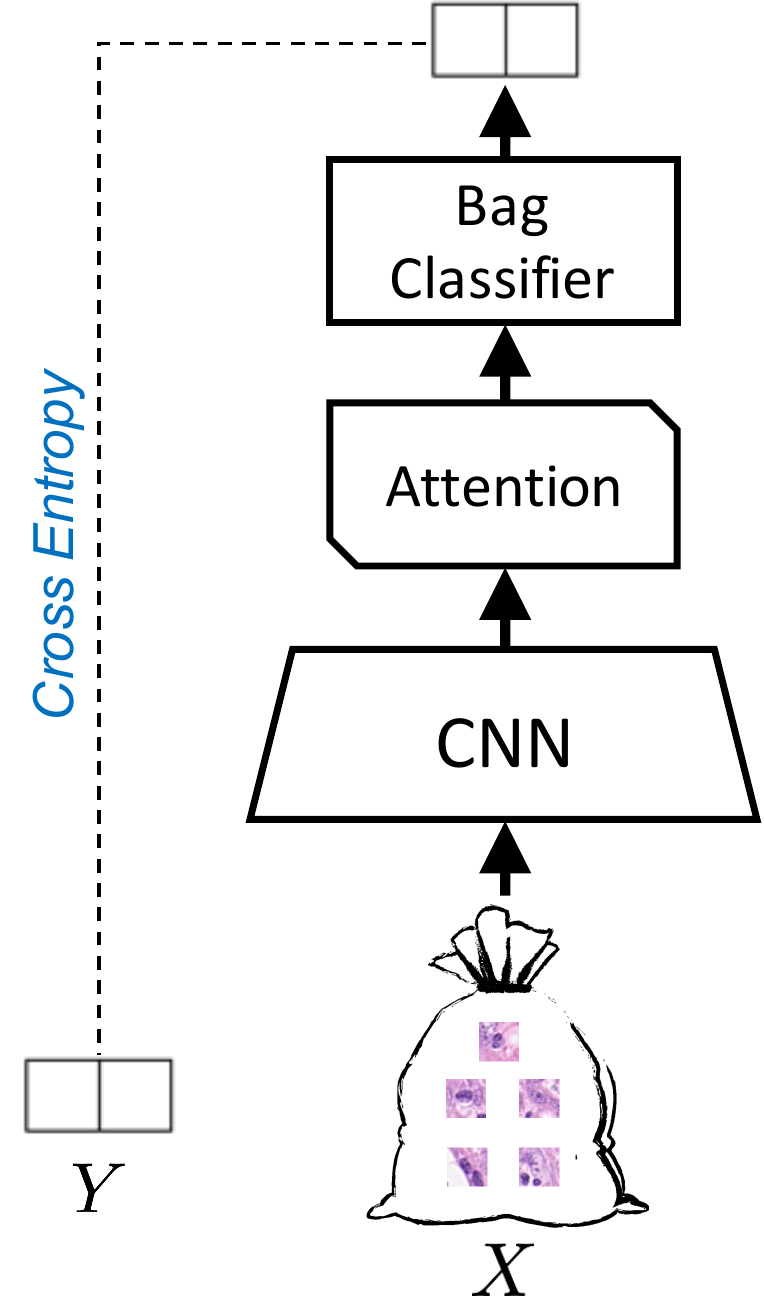}}
	\hspace{0.5in}
	\subfigure[Student model - Instance Label Generator]{\includegraphics[width=0.54\linewidth]{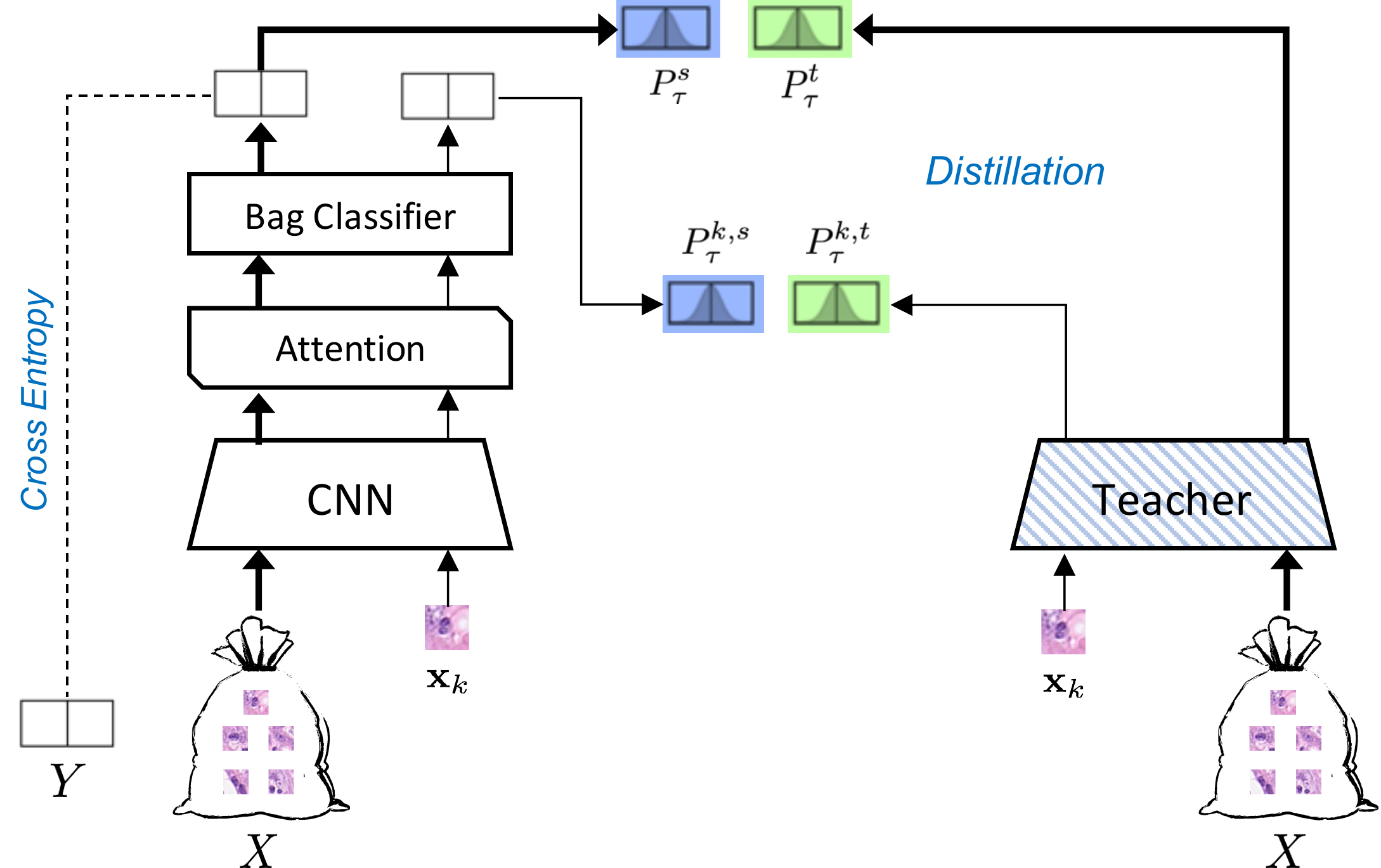}}
	\caption{An illustration of the proposed approach for weakly supervised instance labeling in image classification tasks. Each image is represented as a bag of patches and a teacher model $\mathcal{T}$ is trained to predict bag labels. Subsequently, we build a student model $\mathcal{S}$ that jointly optimizes bag-level prediction with respect to the ground truth, and instance-level prediction based on knowledge distillation from the teacher.}
	\label{fig:approach}
\end{figure*}

Since we assume no access to the instance-level labels, we propose to utilize \textit{knowledge distillation} to leverage the pre-trained MIL model to design an effective instance-level label generator. In general, knowledge distillation (KD) involves transferring from one machine learning model (\textit{teacher}) to another (\textit{student})~\cite{hinton2015distilling}, such that performance on the task under consideration improves, or the number of parameters required for solving a task reduces. In its classical formulation, the teacher is a high-capacity model with high performance, while the student is designed to be more compact. However, several recent works have demonstrated KD to be very effective in transferring to high-capacity student models~\cite{furlanello2018born}. In this work, we view knowledge distillation from a different perspective by identifying what parts of the teacher's expertise (bag-level prediction) can be trusted while training the student model that can make robust predictions for single instances, i.e., bags of size $1$. We introduce a novel formulation that jointly distills from bag-level and instance-level predictions of the teacher to produce highly effective instance label generators, using only weakly supervised data. To the best of our knowledge, this is the first known approach for using distillation to repurpose MIL models for instance labeling. 


Using empirical studies on colon cancer and breast cancer detection from histopathological images, we demonstrate significant improvements to instance-level labeling quality, when compared to the state-of-the-art MIL technique in~\cite{ilse2018attention}. Interestingly, we find that the proposed VAT regularization leads to superior results on these challenging benchmark problems over existing MIL approaches, particularly in terms of instance-level prediction. Furthermore, incorporating our knowledge distillation strategy resulted in additional performance gains. Notably, our method provides improvements of $10\%$ and $37\%$ on the instance labeling accuracy for colon and breast cancer datasets respectively. In summary, we show that distillation can be an effective strategy for automatically generating dense labels, which can make models interpretable for clinical practice, and can support subsequent training of dense segmentation models without the need for dense annotations.

\section{Problem Setup}
Following the formulation of multiple instance learning, we consider each sample to be a bag of instances, $X = \{\mathbf{x}_1, \cdots, \mathbf{x}_K\}$, with no specific dependency or ordering between them. Note that, in our setup, each instance $\mathbf{x}_k$ corresponds to an image patch and the number of instances $K$ can vary between bags. The target variable $Y \in \{0,1\}$ corresponds to the label for a bag, whereas each $y_k \in \{0,1\}$ denotes the label for each of the instance. Since our approach is weakly supervised, we assume there is no access to the instance-level labels during training. The goal is to build a teacher model $\mathcal{T}$ based on MIL to perform bag-level prediction, and subsequently employ knowledge distillation to train an identical student model $\mathcal{S}$ that can produce predictions for both $Y$ and $\{y_k\}$. Since the predictions in MIL should be invariant to the permutations of the instance order in the bags, we consider prediction functions of the following form.
\begin{thm}
A prediction function for a set of instances $X$, $\mathcal{F}(X) \in \mathbb{R}$ is permutation-invariant to the elements in $X$, iff, it can be expressed as follows:
$$\mathcal{F}(X) = g\left(\sum_k f(\mathbf{x}_k)\right),$$where $f$ and $g$ are appropriate transformations. 
\end{thm}Note that, this universal decomposition can be replaced by approximations including the \textit{max} operation in lieu of the summation in the above theorem. Since we employ neural networks to process the raw data, the above theorem can be modified to operate with the latent features $\{\mathbf{h}_k \in \mathbb{R}^d\}$ instead of $\{\mathbf{x}_k\}$, where $\Phi: \mathbf{x}_k \mapsto \mathbf{h}_k$ denotes the mapping learned by the network. In that case, the transformation $f$ is simplified to \textit{identity} and the transformation $g: \mathbb{R}^d \mapsto \mathbb{R}$ maps from the $d-$dimensional latent space to the target $Y$.

\section{Proposed Approach}
In this section, we describe the proposed methodology for building an effective instance label generator using only weakly supervised data. An overview of our approach can be found in Figure \ref{fig:approach}.

\subsection{Teacher Model -- Multiple Instance Learning}
The teacher model is designed to predict the bag-level target $Y$ using the set of instances $X$ using MIL. Our MIL formulation is similar to the one found in~\cite{ilse2018attention}, wherein an embedding-level aggregation strategy is used to perform bag classification. As shown in Figure \ref{fig:approach}(a), each of the instances in the input bag is processed using a convolutional neural network. The resulting features for each of the instances are then aggregated using a permutation-invariant pooling strategy before invoking the classifier module. Though the pooling can be carried out using simple strategies such as \textit{instance average} or \textit{instance max}, they are not directly trainable. Instead, we perform a weighted average of instance features, where the weights are determined using an attention module implemented as a simple neural network. Further, we include a constraint that the weights must sum to $1$, in order to ensure that the aggregation can be applied to bags of varying sizes. 

Denoting the bag of features by $H = \{\mathbf{h}_1, \cdots, \mathbf{h}_K\}$, the aggregation strategy can be formally stated as
\begin{equation}
\mathbf{z} = \sum_k a_k \mathbf{h}_k,
\label{eqn:agg}
\end{equation}where the computation of weights $a_k$ is parameterized using a gated attention mechanism:
\begin{equation}
a_k = \frac{\exp\left(\mathbf{w}^T \left(\text{tanh}(\mathbf{V} \mathbf{h}_k^T) \odot \text{sigm}(\mathbf{U} \mathbf{h}_k^T)  \right)\right)}{\sum_{j=1}^K \exp\left(\mathbf{w}^T \left(\text{tanh}(\mathbf{V} \mathbf{h}_j^T) \odot \text{sigm}(\mathbf{U} \mathbf{h}_j^T)  \right)\right)}.
\label{eqn:att}
\end{equation}Here, $\mathbf{w} \in \mathbb{R}^{L \times 1}$, $\mathbf{U} \in \mathbb{R}^{L \times d}$, $\mathbf{V} \in \mathbb{R}^{L \times d}$ are learnable parameters of the attention module. Note that, the attention computation is posed as a softmax function such that the constraint of weights summing to $1$ is satisfied. Finally, by using a combination of $\text{tanh}$ and $\text{sigm}$ activations enables both the inference of non-linear relations between instances and more importantly removes the troublesome linearity in the $\text{tanh}(.)$ function. Note, the teacher model is trained using the cross entropy loss and optimized to maximize prediction performance on target variable $Y$.

\subsection{Virtual Adversarial Training for MIL}
Though attention based aggregation can be superior to non-trainable pooling strategies, it is more prone to overfitting, particularly when the number of bags $N$, or the number of instances in each bag $K$ are low. This limitation can be particularly severe when we utilize a pre-trained MIL model to produce instance-level labels by inputting bags with a single instance. Hence, we propose to incorporate \textit{virtual adversarial training} based regularization (VAT) to promote robustness with respect to arbitrary perturbations of the bags. VAT strategies have been successfully utilized in supervised and semi-supervised learning formulations~\cite{shu2018dirt}, wherein locally-Lipschitz constraint is incorporated by enforcing classifier consistency around the norm-ball neighborhood of each sample. 

In the context of MIL, we propose to enforce three consistency conditions for the teacher model: (i) $\mathcal{T}$ must produce consistent predictions even when uncorrelated noise instances are included; (ii) $\mathcal{T}$ must produce consistent predictions when each instance in a bag is perturbed around its norm-ball neighborhood; and (iii) the conditional entropy of predictions from $\mathcal{T}$ must be low even when a random subset of instances are arbitrarily excluded from a bag. Hence the overall loss function for the proposed MIL can be written as:
\begin{align}
\nonumber \mathcal{L}(\mathcal{T}) &= \lambda_{c}\mathcal{L}_{c}(Y, \mathcal{T}(X)) + \lambda_{n} \mathcal{L}_n(\mathcal{T}(X), \mathcal{T}(X_n))  \\
& + \lambda_{\delta} \mathcal{L}_{\delta}(\mathcal{T}(X), \mathcal{T}(X_{\delta})) + \lambda_{e} \mathcal{L}_{e}(\mathcal{T}(\bar{X})).
\label{eqn:loss-mil}
\end{align}Here, $X$ is a training bag with $K$ instances. Now, $X_n$ denotes a modified bag with $K_n > K$ instances that includes $K_n - K$ uncorrelated noise instances, i.e.,
$$
X_n =  \{\mathbf{x}^n_k\}_{k = 1}^{K_n}, \text{ where } \mathbf{x}^n_k = \begin{cases}
\mathbf{x}_k, & \text{if}\ k \leq K \\
\tilde{\mathbf{n}}_k, & \text{otherwise}.
\end{cases}
$$Here, $\tilde{\mathbf{n}}_k$ corresponds to an uncorrelated noise instance drawn from a uniform random distribution defined in the same range as the instances in the training data. Next, in eq. (\ref{eqn:loss-mil}), $X_{\delta}$ is obtained by perturbing each instance in $X$ within a norm ball of radius $\delta$:
$$
X_{\delta} =  \{\mathbf{x}^{\delta}_k\}_{k = 1}^{K}, \text{ where } \mathbf{x}^{\delta}_k = \mathbf{x}_k + \tilde{\delta},
$$where $\tilde{\delta}$ denotes a random perturbation within the norm ball around $\mathbf{x}_k$. Finally $\bar{X}$, containing $\bar{K} < K$, instances is obtained by excluding a random subset of instances from $X$:
$$
\bar{X} =  \{\bar{\mathbf{x}}_{k}\}_{k = 1}^{\bar{K}}, \text{ where }, \bar{\mathbf{x}}_k =  \mathbf{x}_{\mathcal{I}(k)},
$$where $\mathcal{I}$ denotes a randomly selected subset of indices of instances from the bag $X$.

While the three loss functions $\mathcal{L}_{c}, \mathcal{L}_{n}$ and $\mathcal{L}_{\delta}$ are implemented using the binary cross entropy function, the final term $\mathcal{L}_{e}$ uses the conditional entropy of the output probabilities. Note, minimizing the conditional entropy ensures that the output probabilities from the model is concentrated around one of the classes. The choice of the hyper-parameters for this optimization are discussed in the experiments section.

\vspace{0.1in}

\begin{figure}[t]
	\centering
	\subfigure[10 instances per bag on average]{\includegraphics[width=0.99\linewidth]{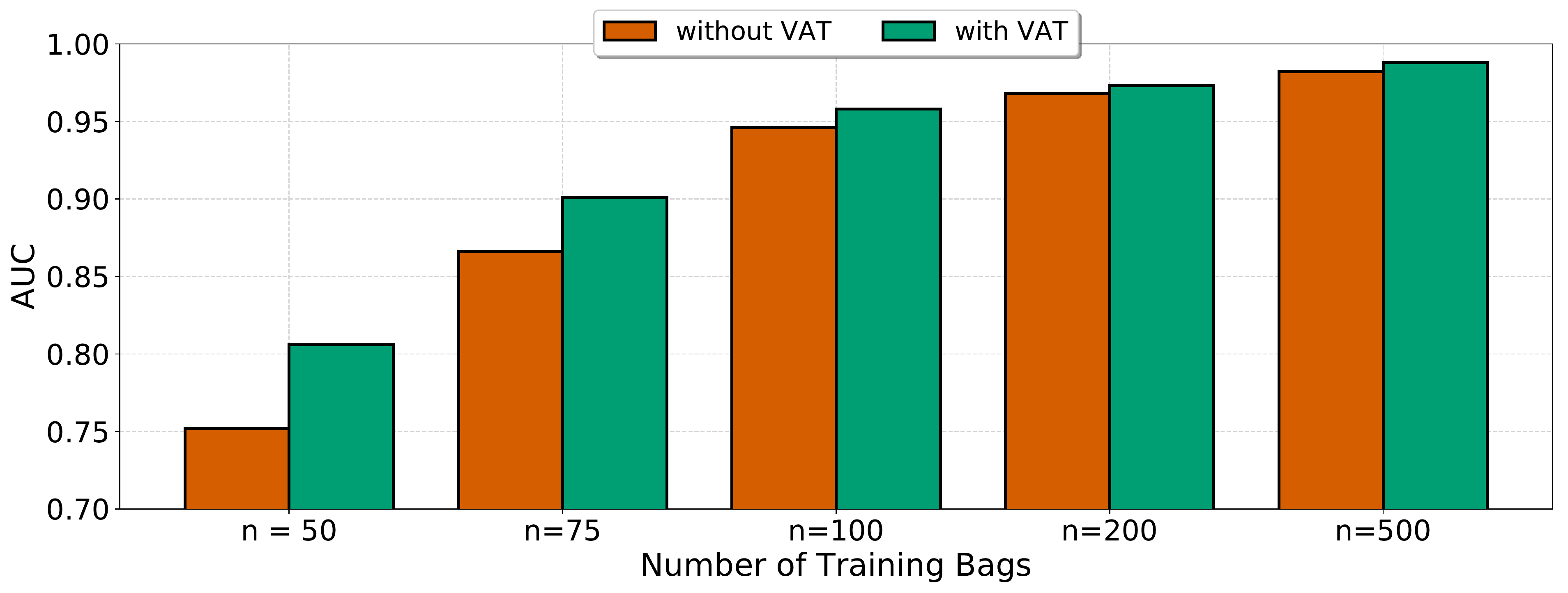}}
	\subfigure[20 instances per bag on average]{\includegraphics[width=0.99\linewidth]{./figures/mnist_10instperbag.pdf}}
	\caption{Effect of the proposed virtual adversarial training (VAT) on the performance of the teacher model. This experiment is carried out with the MNIST digits dataset, with varying number of bags $N$ and average number of instances per bag $K$.}
	\label{fig:mnist}
\end{figure}
\noindent \textbf{Effect of VAT.} In order to understand the usefulness of including VAT into the MIL training process, we setup an empirical study with the MNIST handwritten digits data, similar to~\cite{ilse2018attention}. A bag is composed of a random number of images from the MNIST dataset, and is assigned a positive label if it contains one or more images of digit $9$. The number of instances in a bag is Gaussian-distributed with mean $K$ (closest integer value is taken) and variance $5$. We study the effect of VAT by varying the number of bags $N = 50, 75, 100, 200, 500$ and the average number of instances per bag $K = 10, 20$ respectively. We used a fixed set of $1000$ bags for validating the resulting models from each of the cases. All experiments were run with the LeNet5 architecture, trained using the RMSprop optimizer and the area under the ROC (receiver operating characteristic) curve is used as the evaluation metric. Figure \ref{fig:mnist} shows the bag-level prediction performance on the validation set for the different cases. As it can be observed, the proposed VAT regularization leads to significant performance improvements for varying bag sizes, particularly when $N$ is low, thus resulting in highly consistent predictive models.

\subsection{Student Model -- Instance Label Generation}
Since we do not have access to ground truth labels for training the instance-level classifier, we propose to leverage knowledge from the teacher model, i.e. bag-level classifier, using a knowledge distillation approach. More specifically, we build a student model $\mathcal{S}$ that has an architecture similar to the teacher, but with the goal of improving its prediction quality for the instances. An interesting aspect of the proposed formulation is that we do not need require a separate predictor for the instance-level, since a single instance can be represented as a bag of size $1$. Following our notations defined earlier, for a given bag comprising the instances $X$, we denote the features obtained using the teacher and student models as $H^t$ and $H^s$ respectively. Given the latent representations both the teacher and student models, we perform aggregation using the attention module and employ an output \textit{linear} layer to produce the logits, followed by the \textit{softmax} function to obtain the probability of assignment for each of the classes.

\begin{figure}
	\centering
	\includegraphics[width=0.9\linewidth]{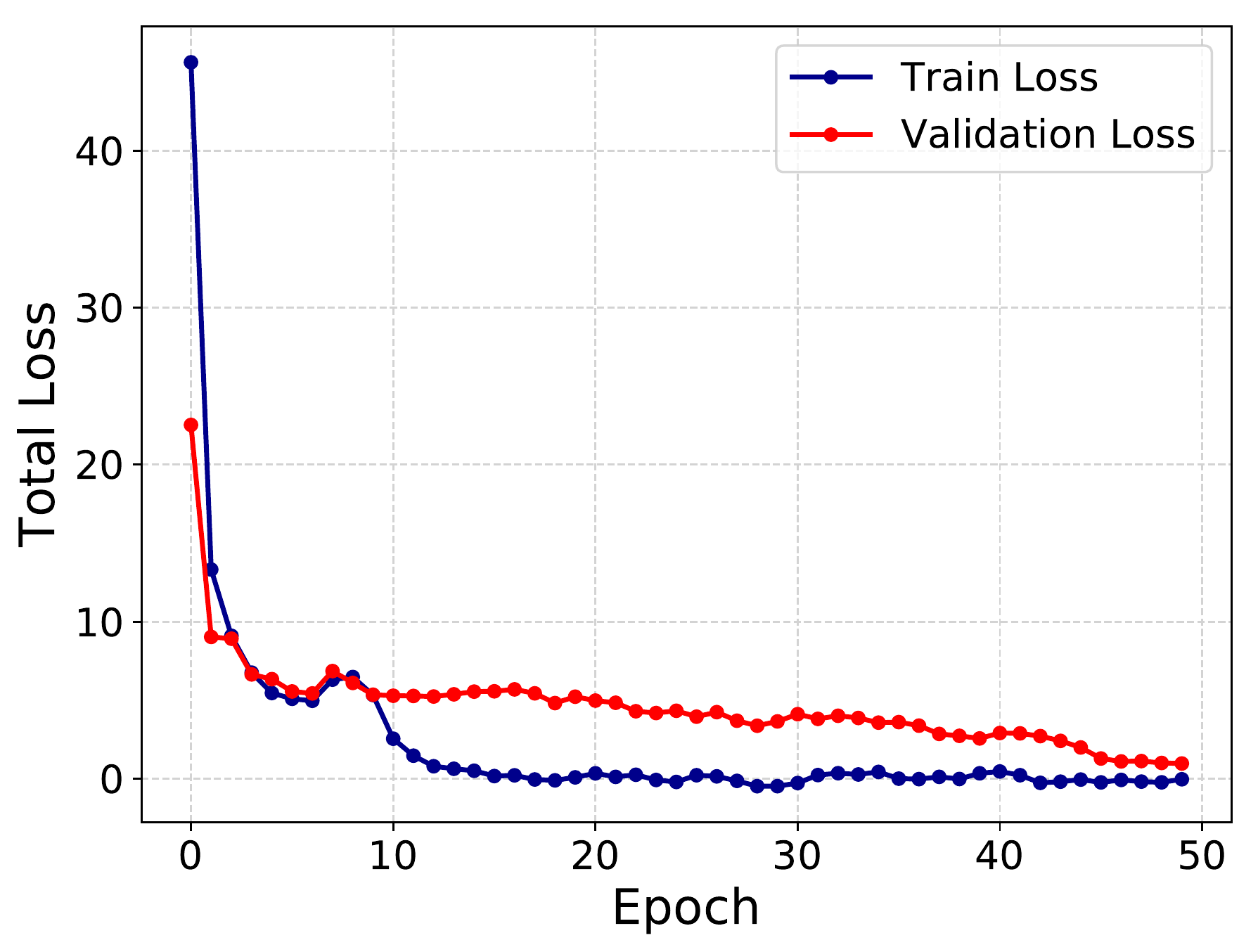}
	\caption{Colon cancer dataset - Convergence behavior of the student model training process. Note, the teacher was pre-trained using MIL along with the proposed VAT regularization. The total loss showed here is obtained using the objective in eq. (\ref{eqn:obj})}.
	\label{fig:losses}
\end{figure}

\begin{figure}
	\centering
	\subfigure[Colon cancer dataset]{\includegraphics[width=0.9\linewidth]{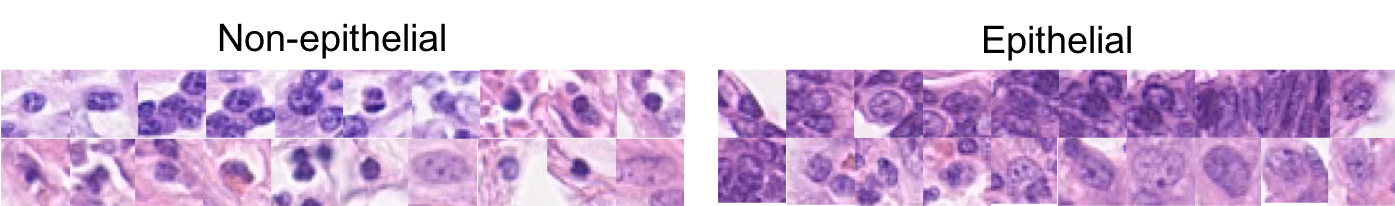}}
	\subfigure[Breast cancer dataset]{\includegraphics[width=0.9\linewidth]{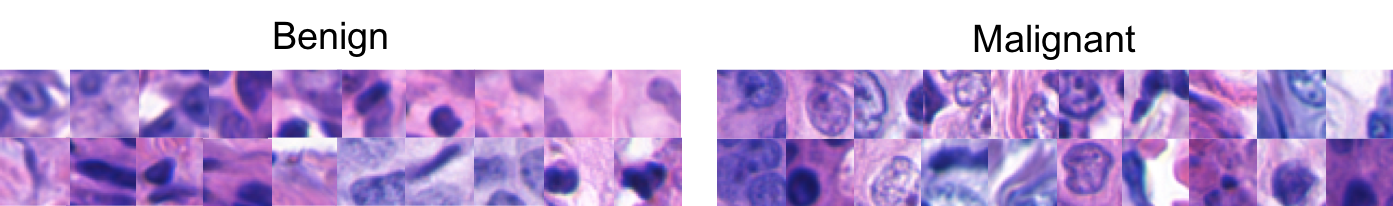}}
	\caption{Illustration of instances from the two histopathological datasets used in our experiments.}
	\label{fig:examples}
	\vspace{-0.1in}
\end{figure}

\begin{table*}[t]
	\centering
	\caption{Instance-level label prediction performance for the colon cancer dataset. We show results from $5-$fold cross validation obtained using the baseline MIL~\cite{ilse2018attention}, the proposed teacher and student models. The average performance across the $5$ folds is also included for each case.}
	\label{tab:coloncancer}
	\renewcommand*{\arraystretch}{1.4}
	\begin{tabular}{c|c|c|c|c|c|c|c|c|c}
		\hline
		
		\multirow{2}{*}{\textbf{Fold}} & \multicolumn{3}{c|}{\textbf{Accuracy}} & \multicolumn{3}{c|}{\textbf{F1 Score}} & \multicolumn{3}{c}{\textbf{AUROC}} \\ 
		\cline{2-10}
		& \textbf{MIL~\cite{ilse2018attention}} & \begin{tabular}[c]{@{}c@{}}\textbf{Teacher}\end{tabular} & \begin{tabular}[c]{@{}c@{}}\textbf{Student}\end{tabular} & \textbf{MIL~\cite{ilse2018attention}} & \begin{tabular}[c]{@{}c@{}}\textbf{Teacher}\end{tabular} & \begin{tabular}[c]{@{}c@{}}\textbf{Student}\end{tabular} & \textbf{MIL~\cite{ilse2018attention}} & \begin{tabular}[c]{@{}c@{}}\textbf{Teacher}\end{tabular} & \begin{tabular}[c]{@{}c@{}}\textbf{Student}\end{tabular} \\ \hline \hline
		0 & 0.80 & 0.83 & \textbf{0.88} & 0.47 & 0.61 & \textbf{0.75} & 0.78 & 0.82 & \textbf{0.87} \\ \hline
		1 & 0.79 & 0.84 & \textbf{0.90} & 0.71 & 0.79 & \textbf{0.81} & 0.89 & \textbf{0.92} & \textbf{0.92} \\ \hline
		2 & 0.69 & 0.79 & \textbf{0.83} & 0.48 & 0.64 & \textbf{0.75} & 0.87 & 0.90 & \textbf{0.91} \\ \hline
		3 & 0.84 & 0.90 & \textbf{0.92} & 0.68 & 0.87 & \textbf{0.90} & 0.89 & \textbf{0.92} & \textbf{0.92} \\ \hline
		4 & 0.83 & 0.88 & \textbf{0.90} & 0.71 & 0.75 & \textbf{0.78} & 0.88 & 0.91 & \textbf{0.92} \\ \hline
		\hline
		\rowcolor{gray!30}
		\textbf{\textit{Average}} & \textit{0.79} & \textit{0.85} & \textbf{\textit{0.89}} & \textit{0.61} & \textit{0.73} & \textbf{\textit{0.80}} & \textit{0.86} & \textit{0.89} & \textbf{\textit{0.91}} \\ \hline
	\end{tabular}
\end{table*}

\begin{table*}[t]
	\centering
	\caption{Instance-level label prediction performance for the breast cancer dataset. We show results from $5-$fold cross validation obtained using the baseline MIL~\cite{ilse2018attention}, the proposed teacher and student models. The average performance across the $5$ folds is also included for each case.}
	\label{tab:breastcancer}
	\renewcommand*{\arraystretch}{1.4}
	\begin{tabular}{c|c|c|c|c|c|c|c|c|c}
		\hline
		\multirow{2}{*}{\textbf{Fold}} & \multicolumn{3}{c|}{\textbf{Accuracy}} & \multicolumn{3}{c|}{\textbf{F1 Score}} & \multicolumn{3}{c}{\textbf{AUROC}} \\ 
		\cline{2-10}
		& \textbf{MIL~\cite{ilse2018attention}} & \begin{tabular}[c]{@{}c@{}}\textbf{Teacher}\end{tabular} & \begin{tabular}[c]{@{}c@{}}\textbf{Student}\end{tabular} & \textbf{MIL~\cite{ilse2018attention}} & \begin{tabular}[c]{@{}c@{}}\textbf{Teacher}\end{tabular} & \begin{tabular}[c]{@{}c@{}}\textbf{Student}\end{tabular} & \textbf{MIL~\cite{ilse2018attention}} & \begin{tabular}[c]{@{}c@{}}\textbf{Teacher}\end{tabular} & \begin{tabular}[c]{@{}c@{}}\textbf{Student}\end{tabular} \\ \hline \hline
		0 & 0.29 & 0.59 & \textbf{0.64} & 0.19 & 0.47 & \textbf{0.53} & 0.51 & 0.71 & \textbf{0.77} \\ \hline
		1 & 0.21 & 0.54 &\textbf{0.60} & 0.23 & 0.51 & \textbf{0.55} & 0.61 & 0.72 & \textbf{0.76} \\ \hline
		2 & 0.32 & 0.63 & \textbf{0.66} & 0.29 & 0.62 & \textbf{0.64} & 0.63 & 0.77 & \textbf{0.78} \\ \hline
		3 & 0.27 & 0.58 & \textbf{0.65} & 0.31 & 0.63 & \textbf{0.68} & 0.67 & 0.73 & \textbf{0.74} \\ \hline
		4 & 0.28 & 0.61 & \textbf{0.63} & 0.27 & 0.58 & \textbf{0.60} & 0.59 & 0.64 & \textbf{0.70} \\ \hline
		\hline
		\rowcolor{gray!30}
		\textbf{\textit{Average}} & \textit{0.27} & \textit{0.59} & \textbf{\textit{0.64}} & \textit{0.26} & \textit{0.56} & \textbf{\textit{0.60}} & \textit{0.60} & \textit{0.71} &\textbf{ \textit{0.75}} \\ \hline
	\end{tabular}
\end{table*}

Formally, for an input $X$, the teacher model produces the output $P^t = \texttt{softmax}(z^t)$, where the aggregated representation $z^t$ is obtained using eq. (\ref{eqn:agg}). Similarly, the student $\mathcal{S}$ produces the output $P^s = \texttt{softmax}(z^s)$. In the original formulation of knowledge distillation~\cite{hinton2015distilling}, the student is trained such that its output $P^s$ is similar to the teacher’s output $P^t$, as well as to the true labels, in order to improve the bag-level classification performance. In practice, the output probability of the teacher is smoothed with a temperature $\tau$ to soften the estimate, and provide additional information about the challenges during training.
\begin{equation}
P^t_{\tau} = \texttt{softmax} \left(\frac{z^t}{\tau}\right), P^s_{\tau} = \texttt{softmax} \left(\frac{z^s}{\tau}\right).
\label{eqn:kd}
\end{equation}However, in our setup, the goal is not improve the bag-level classification, but instead the instance-level prediction, without accessing ground truth labels entirely. To this end, we propose this optimization objective:
\begin{align}
\nonumber \mathcal{L}(\mathcal{S}) = &\gamma_{c}\mathcal{L}_{c}(Y, P^s) + \gamma_{b}\mathcal{H}(P^t_{\tau},P^s_{\tau})
\\
& + \gamma_{i} \sum_{k = 1}^K \mathcal{H}(P^{k, t}_{\tau},P^{k, s}_{\tau}) + \gamma_e \sum_{k =1}^K L_e(P^{k, s}).
\label{eqn:obj}
\end{align}Here, $P^{k, s}$ denotes the output from $\mathcal{S}$ for the $k^{\text{th}}$ instance in $X$, and $\mathcal{H}$ is the KL-divergence between two Bernoulli variables:
\begin{equation}
\mathcal{H}(p,q) = p \log \frac{p}{q} + (1-p) \log \frac{(1-p)}{(1-q)}.
\end{equation}In the above objective, the first term denotes the binary cross entropy between the predicted bag-level labels and the ground truth, while the second term performs distillation from $\mathcal{T}$ to $\mathcal{S}$ using the softened softmax probabilities. The last two terms are based on the instance-level predictions obtained by passing each instance ($\mathbf{x}_k$) independently through the teacher and the student models. Similar to the formulation in eq. (\ref{eqn:loss-mil}), $L_e$ denotes the conditional entropy. Through this controlled knowledge distillation at bag and instance levels, we show that our approach achieves significant gains in instance-level prediction even without accessing supervisory labels.


\section{Experiments}
In this section, we describe the two applications considered, discuss the experiment setup, and present a detailed performance evaluation. All experiments were carried based on a standard convolutional neural network comprising a stack of convolution layers with ReLU activation followed by max pooling, and $3$ fully connected layers, while the classifier contains a single fully connected layer. For comparison, we consider the state-of-the-art baseline~\cite{ilse2018attention}, that employs attention-based aggregation to perform MIL\footnote{Implementation for attention-based MIL was obtained from a. \url{https://github.com/utayao/Atten_Deep_MIL} and b. \url{https://github.com/AMLab-Amsterdam/AttentionDeepMIL}.}. All models were implemented using PyTorch~\cite{paszke2017automatic}.  

\subsection{Dataset Description}
We tested our algorithm on cancer histopathology datasets of the breast and colon. Histopathology images are especially relevant because the nucleus identification of diseased cells is a highly specialized and time-consuming task. Consequently, it would be hugely beneficial if the relevant instances of tumorous patches could be identified accurately, which would in turn greatly reduce the burden on clinicians. 

\subsubsection{Color Cancer Dataset}
This dataset is comprised of $100$ Hematoxylin and eosin stain (H\&E) histopathology ($500 \times 500$ pixels) images of colorectal cancer~\cite{sirinukunwattana2016locality}. Out of the 29,756 nuclei centers identified, 22,444 nuclei were given an associated class label, i.e. epithelial, inflammatory, fibroblast, and miscellaneous. To simplify our problem for the multiple instance learning we chose to focus on the binary problem of detecting the epithelial cell nucleus. This is motivated by the findings in~\cite{ricci2007identification} that mutations expressed in the epithelial cells of the colon are often the starting point for cancerous mutation. In this case, each whole slide image (bag) was divided into $27 \times 27$ image patches (instances) (See Fig.~\ref{fig:examples} for examples). Every bag containing epithelial cells were marked as positive. Further, the ground truth labels for instances, i.e. patches, were formed by studying if the instance contained epithelial nuclei. Note, these instance labels were used only for evaluation and not utilized during training. 

\subsubsection{Breast Cancer Dataset}
This dataset consists of $58$ H\&E stained histopathology images ($896 \times 768$ pixels)~\cite{Drelie08-298} with each of the cells marked as benign or malignant. Further, each whole slide cell image (bag) is also assigned a  positive label if it contained malignant cancer cells (instances). Similar to the previous case, the images were divided into $32 \times 32$ patches to form the instances (See Fig.~\ref{fig:examples} for examples of both). Note, with those instances with 75\% or more of white pixels were considered as background and discarded from our study. 

Both experiments were carried out on a train/valid/test split of $70/10/20$ with $5-$fold cross validation.The number of convolution layers in the feature extractor part of the network was fixed at $2$ for the colon cancer dataset and at $5$ for the breast cancer dataset. We used the Adam optimizer with learning rate set to $1e-4$ and used batch size $1$. The hyper-parameters for the student model training in  eq. (\ref{eqn:obj}) were set at the following values for both cases: $\gamma_{c} = 0.3, \gamma_{b} = 0.5, \gamma_{i} = 0.5, \gamma_e = 0.1$. On the other hand the hyper-parameters for teacher model training in eq. (\ref{eqn:loss-mil}) were tuned specifically for the two datasets. This was because the breast cancer dataset presented a severe class imbalance (benign vs malignant) at the instance-level. More specifically, for the colon cancer dataset, we used the values $\lambda_{c} = 1.0 , \lambda_{n} = 0.5,  \lambda_{\delta} = 0.3, \lambda_{e} = 0.3$, whereas for the latter $\lambda_{c} = 0.8 , \lambda_{n} = 0.8,  \lambda_{\delta} = 0.3, \lambda_{e} = 0.3$. Furthermore, for the breast cancer case, it was beneficial to include multiple random realizations (set to $5$) of $\bar{X}$ for the VAT regularization in eq. (\ref{eqn:loss-mil}). In other words the loss term $L_e$ was constructed by averaging the conditional entropy obtained using bags containing different subsets of instances.

\subsection{Results}
Tables \ref{tab:coloncancer} and \ref{tab:breastcancer} present the cross validated results for the two datasets on the instance-level label generation. As mentioned earlier, we report the results for the state-of-the-art MIL method in~\cite{ilse2018attention}, the teacher and student models from the proposed approach. For evaluation, we adopted the widely adopted metrics, accuracy, F1 score and area under the receiver operating curves (AUROC). The first striking observation is that the proposed approach consistently produces significant performance gains, in terms of all three metrics, over the baseline method. For example, on the colon cancer dataset, our approach provides improvements of $10\%$ in accuracy and $0.19$ in F1 score. More interestingly, with the challenging breast cancer dataset, we find that the baseline MIL performs poorly than even random predictions. In contrast, the proposed strategies lead to more reliable predictions, producing a boost of $37\%$ in the prediction accuracy score. This clearly evidences the effectiveness of the VAT regularization and our novel distillation formulation, in the context of multiple instance learning, towards generating dense labels from weak supervision.

%


\section{Conclusions}
In this paper we presented a method for producing instance-level labels with only weak supervision, i.e. image-level labels, in medical image analysis. Our method relies on using a novel virtual adversarial training regularization to MIL and repurposing a pre-trained MIL model for instance classification. Through empirical validation on two very challenging histopathology cancers of the colon and the breast, we showed that the proposed method consistently outperformed the state-of-the-art MIL. This presents a huge opportunity to save exhaustive efforts in annotating clinical data, and to more importantly enable weakly supervised data augmentation for data-driven inferencing. Future work will include extending this methodology to multi-class predictions, and developing scalable techniques to produce pixel-level segmentation.


\small

\bibliographystyle{IEEEbib}
\bibliography{main}

\end{document}